\documentclass[twocolumn,9pt]{article}
\usepackage{extsizes}
\usepackage{graphicx} 
\usepackage{geometry}

\usepackage[compact]{titlesec}
\usepackage{tgpagella}

\usepackage{gensymb}
\usepackage{hyperref}
\usepackage{lipsum}
\usepackage{float}
\usepackage{dblfloatfix}
\usepackage{mathtools}
\usepackage{sectsty}
\usepackage{subcaption}
\usepackage[utf8]{inputenc}
\usepackage{enumitem}
\usepackage{fancyhdr}
\usepackage{multirow}
\usepackage{multirow}
\usepackage{graphicx}
\usepackage{verbatim}
\usepackage{booktabs}
\usepackage{cite}
\usepackage{sectsty}

\title{\textbf{IIT Bombay Racing Driverless:\\ Autonomous Driving Stack for Formula Student AI}}
\author{
\begin{tabular}{c}
Yash Rampuria$^{1}$, Deep Boliya$^{2}$, Shreyash Gupta$^{1}$, Gopalan Iyengar$^{1}$, Ayush Rohilla$^{3}$, Mohak Vyas$^{4}$, Chaitanya Langde$^{4}$,\\ Mehul Vijay Chanda$^{5}$, Ronak Gautam Matai$^{6}$, Kothapalli Namitha$^{7}$,  Ajinkya Pawar$^{3}$,  Bhaskar Biswas$^{5}$,\\Nakul Agarwal$^{4}$,
Rajit Khandelwal$^{3}$, Rohan Kumar$^{1}$, Shubham Agarwal$^{1}$, Vishwam Patel$^{4}$,\\ Abhimanyu Singh  Rathore$^{5}$, Amna Rahman$^{8}$, Ayush Mishra$^{1}$, Yash Tangri$^{1}$
\footnote{
$^{1}$Department of Mechanical Engineering, IIT Bombay\\   
$^{2}$Department of Electrical Engineering, IIT Bombay\\
$^{3}$Department of Metallurgical Engineering and Materials Science, IIT Bombay\\
$^{4}$Department of Civil Engineering, IIT Bombay\\
$^{5}$Department of Physics, IIT Bombay\\
$^{6}$Department of Energy Science and Engineering, IIT Bombay\\
$^{7}$Department of Computer Science and Engineering, IIT Bombay\\
$^{8}$Department of Chemical Engineering, IIT Bombay
}
\end{tabular}
}
\date{
\begin{tabular}{c}
Indian Institute of Technology Bombay, Mumbai, India
\end{tabular}
}

\begin{document}
\pagenumbering{gobble} 
\setlength{\intextsep}{2pt}
\setlength{\textfloatsep}{7pt}
\setlength{\abovecaptionskip}{4pt}
\setlength{\belowcaptionskip}{0pt}
\urlstyle{same}

\sectionfont{\LARGE}  
\subsectionfont{\large}  
\subsubsectionfont{\normalsize}  

\rhead{\thepage}
\cfoot{}
\fancyfoot{}
\renewcommand{\headrulewidth}{0pt}

\maketitle
\textbf{\textit{Abstract-} This work presents the design and development of IIT Bombay Racing's Formula Student style autonomous racecar algorithm capable of running at the racing events of Formula Student-AI, held in the UK. The car employs a cutting-edge sensor suite of the compute unit NVIDIA Jetson Orin AGX, 2 ZED2i stereo cameras, 1 Velodyne Puck VLP16 LiDAR and SBG Systems Ellipse N GNSS/INS IMU. It features deep learning algorithms and control systems to navigate complex tracks and execute maneuvers without any human intervention. The design process involved extensive simulations and testing to optimize the vehicle's performance and ensure its safety. The algorithms have been tested on a small scale, in-house manufactured 4-wheeled robot and on simulation software. The results obtained for testing various algorithms in perception, simultaneous localization and mapping, path planning and controls have been detailed.
}~

\textbf{\textit{Index Terms-} Autonomous Vehicle, Perception, Simultaneous Localization and Mapping, Path Planning and Controls, Formula Student}

\section{Introduction}
\begin{figure*}
    \centering
    \includegraphics[width = \linewidth]{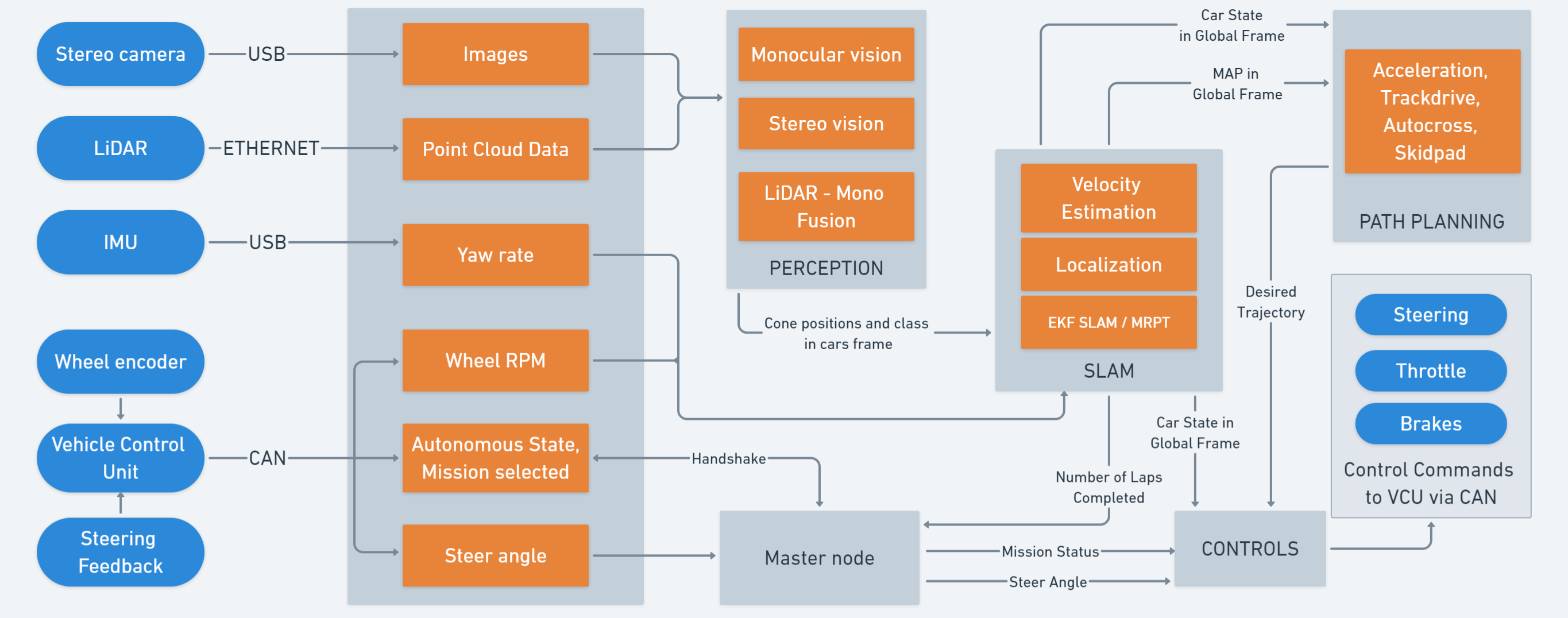}
    \captionsetup{justification=centering} \caption{\textit{High level autonomous system architecture}}
    \label{fig:high_level_software}
\end{figure*}

Formula Student AI is an international racing competition held in the UK. Teams build software systems and deploy them on racecars to race autonomously in different dynamic events: Autocross, Trackdrive, Skidpad and Acceleration, with the race track being defined by cones of classes: blue, yellow, small orange and big orange. Teams take part in Dynamic Driving Task (DDT) or Automated Driving System (ADS) entries as detailed in \cite{b1}. Teams are required to develop only the software stack to race autonomously in the DDT entry and the Autonomous Driving Systems – Dedicated Vehicle (ADS-DV) is provided by IMechE. As of the date of this work, IIT Bombay Racing has built software systems to take part in the DDT entry.

The autonomous system design comprises these major subsystems: Perception, Simultaneous Localization and Mapping (SLAM) and Path Planning and Controls (PPC). Perception uses sensors (Cameras and LiDAR) to detect and understand the environment of the vehicle, in this context, to deduce the spatial coordinates (in the car's frame) and colour of the track landmarks - cones of various sizes and colours as detailed in \cite{b2}. SLAM then uses perception data in the car frame, alongside the odometry estimate of its own position and heading, to build a ground-frame map of the track, and localize itself in said map. This map information along with odometry data are leveraged by the PPC subsystem to determine the best path for the vehicle without hitting cones and staying on track, utilizing advanced control algorithms to calculate the optimal throttle, brake and steering control signals. Interface between the hardware and software is done using ROS2 \& CAN communication while maintaining parallel flow of information between various subsystems. The high level diagram explaining the software structure is given in Fig. \ref{fig:high_level_software}.

The present iteration of the autonomous system has been implemented in pursuit of reliability to make the IMechE Autonomous Driving Systems – Dedicated Vehicle (ADS-DV) operation as safe as possible, deterministic and accurate. Edge and corner cases were thoroughly predicted and tested for. To this end, the following sections detail the components, software development and  results of the algorithms which the final autonomous software system contains.

\section{Perception}

The Perception subsystem is responsible for understanding the environment around the vehicle. This involves classifying detected cones into various classes based on colour and size and also estimating their position (as range and bearing of each cone landmark) relative to the vehicle's frame of reference. A LiDAR and two stereo cameras are utilized for the same. The subsystem pipeline is built with the overall aim of achieving its task with maximum accuracy and range and lowest latency.

\begin{figure}[h]
    \centering
    \includegraphics[width = \linewidth]{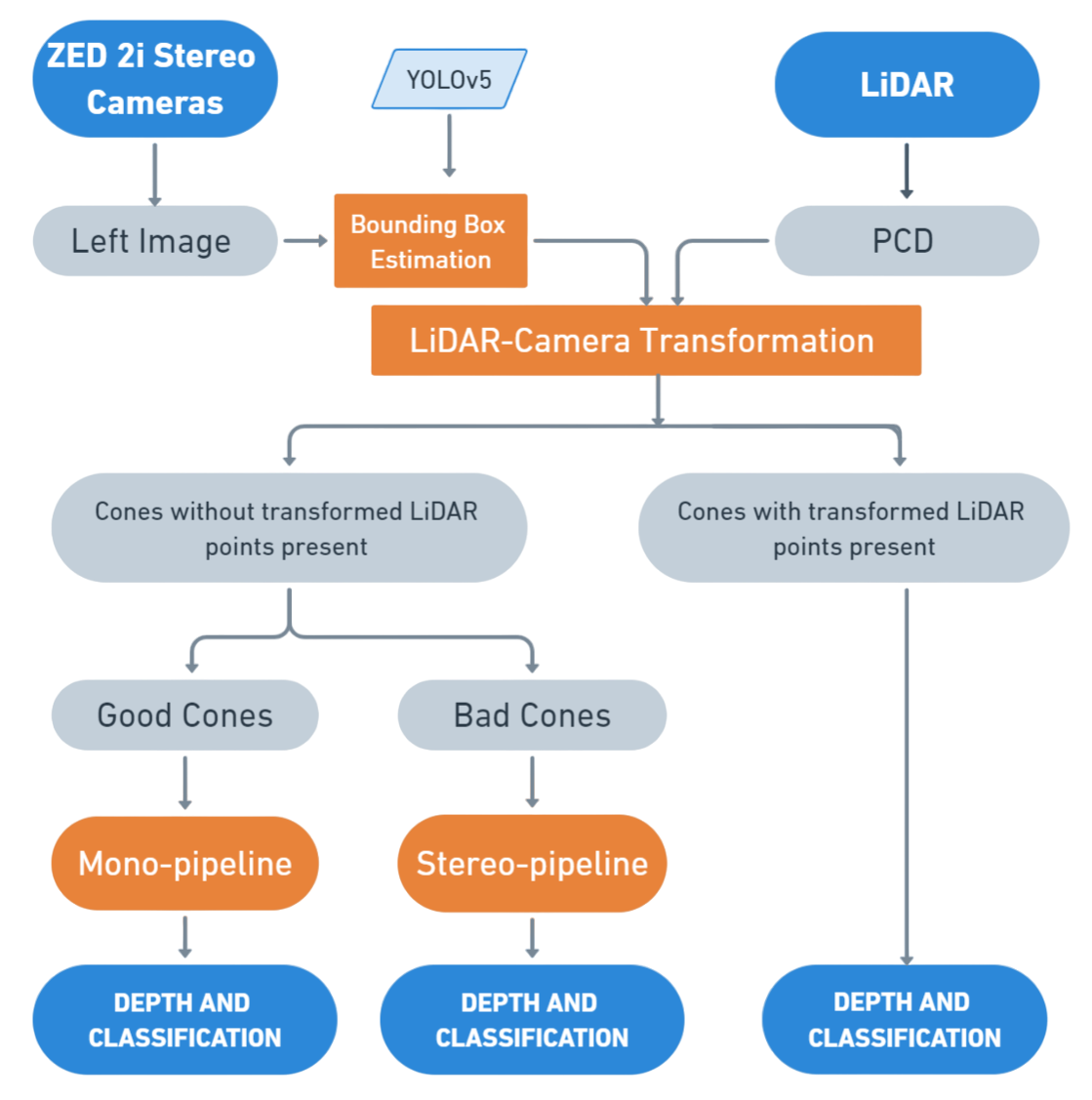}
    \captionsetup{justification=centering} \caption{\textit{The perception subsystem and parallel use of 3 different pipelines}}
    \label{fig:perception_overall}
\end{figure}

With this objective in mind, a three tier integrated approach has been developed which fuses sensor data for various cases. To the best of the authors' knowledge, the combined approach is novel for the purpose of Formula Student AI. YOLOv5 (Section \ref{yolo}) is used to detect cones in images captured. The vertical angular resolution of the LiDAR (Section \ref{hardware}) is 2\degree with 16 channels from -15\degree to +15\degree and thus certain cones (at distance $>$ around 10m) may not be perceived by it. For cones perceived by the LiDAR, the LiDAR-Camera fusion pipeline is used for depth estimation. For undetected cones, a purely camera based depth estimation approach is used. For "good cones" - cones which are upright and fully visible, the monocular pipeline (Section \ref{mono}) is used and for "bad cones", a stereo vision pipeline (Section \ref{stereo}) is used. This 3-tiered approach has been developed by IIT Bombay Racing to leverage the advantages of different depth estimation algorithms and offset their disadvantages - high accuracy and low range of LiDAR, higher range and moderate accuracy of Monocular pipeline and high robustness to cone condition (fallen, blocked, not fully visible, etc.) but lower accuracy of the stereo pipeline. This is depicted in Fig. \ref{fig:perception_overall}. Results are in Table \ref{table:1}.

\subsection{Bounding Box Estimation}
\label{yolo}
The task of detecting the cones in the image and classifying them into categories (color and size) is achieved using the object detection model YOLOv5 \cite{b3}. In 2023 (the previous Formula Student season), YOLOv5s was trained on the FSOCO Dataset \cite{b11}, to which IIT Bombay Racing Driverless was also a contributor. A new dataset was created and trained on this year with images of around 3.5k cones collected during the test runs on the ADS-DV via the ZED2i stereo-camera. These images were used for training the model. Training on the new dataset yielded detection mAP (mean average precision) 0.911. Detection was improved further and mAP increased to 0.985 by using better annotations and augmentation techniques. Blur was added to some images to model motion blur and exposure was varied from -25\% to +25\% to model various illumination conditions. Occlusion modelling for the cones was done using black opaque boxes drawn in front of some cones to simulate cones hidden by other cones.

\begin{figure}
    \centering
    \includegraphics[width = \linewidth, height = 25mm]{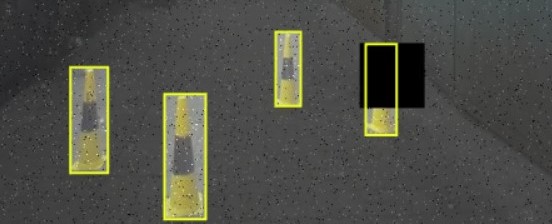}
    \captionsetup{justification=centering} \caption{\textit{Inference by YOLOv5s. Cones are robustly detected even in presence of noise and blockages}}
    \label{fig:yolo}
\end{figure}

YOLOv5 is run on the \textit{left image} yielding bounding boxes on all cones detected. Fig. \ref{fig:yolo} shows the robustness of the new model detection, as it detects even half covered cones on the new noisy dataset with opaque blocks.

\subsection{LiDAR - Camera Fusion Pipeline}
\label{fusion_pipeline}
 
This approach involves processing the vision data separately to ascertain landmark properties (class: colour and size) and using the LiDAR point cloud data (PCD) to determine their 3D location. The steps involved are detailed below.

\begin{figure}[h]
  \centering
    \begin{subfigure}{0.8\linewidth}
    \includegraphics[height=80pt, width=\linewidth]{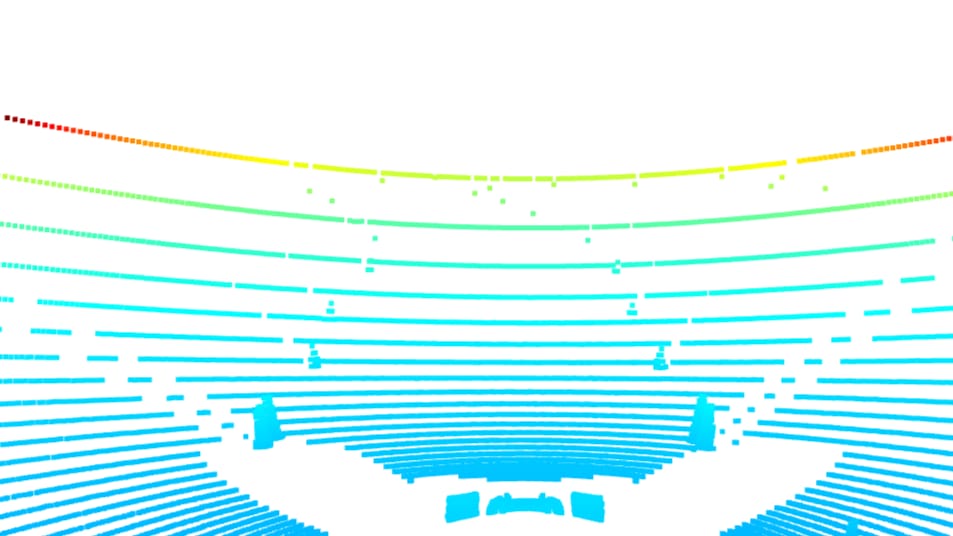}
    \captionsetup{justification=centering} \caption{LiDAR pointcloud before ground removal}
  \end{subfigure}
  \begin{subfigure}{0.7\linewidth}
    \includegraphics[height=90pt, width=\linewidth]{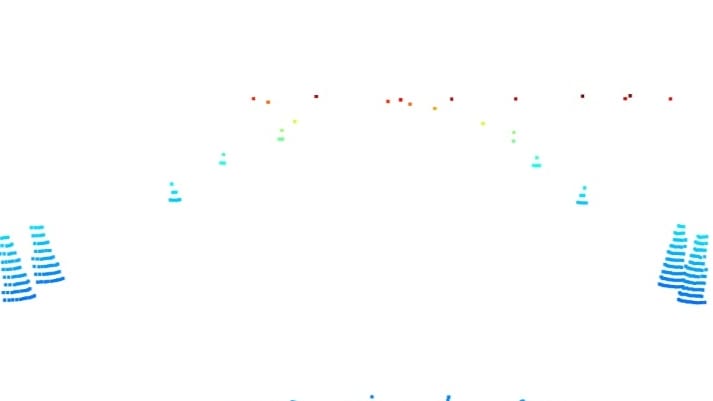}
    \captionsetup{justification=centering} \caption{(Zoomed in) pointcloud after ground removal and clustering}
  \end{subfigure}
  
  \captionsetup{justification=centering} \caption{\textit{Comparision of LiDAR Point Cloud before and after clustering and ground removal. It can be seen that only points on the cones remain}}
  \label{fig:clustering}
  
\end{figure}

\noindent
\textbf{Visual Object Detection}
YOLOv5 is used to detect the bounding boxes in the camera image to make the vision-to-pointcloud correspondence sparse. This provides the necessary class of the cone.
\\
\newline
\textbf{Ground Plane Removal from the LiDAR Point Cloud}
RANSAC (Random Sample Consensus) \cite{b12}, a statistical distribution fitting algorithm is used on the LiDAR point cloud data to segment the ground plane out of the data. It iteratively fits models to subsets of data, identifying points that best fit the model and removing outliers. It is used to isolate the ground plane from the point cloud data.
 The remaining components of the point cloud consist of 3D coordinate data for all cones, and the remaining environment. It is assumed that the ground/track is relatively flat.

\noindent
\\
\textbf{Clustering}
DBSCAN (Density-Based Spatial Clustering of Applications with Noise) \cite{b13} is used to cluster cone data points. Points are grouped based on their density, distinguishing dense regions as clusters while identifying outliers as noise. Objects of varying point densities can be robustly clustered. Cones are specifically retained by setting a height threshold on clusters (as the original height of cones is known beforehand). The result of clustering is point cloud data with only cone clusters remaining. These steps are considered necessary to reduce the number of data points to be projected onto the image frame and also to remove background points which may be erroneously projected onto cones. The original PCD and clustered PCD is showin in Fig. \ref{fig:clustering}.

\vspace{-4pt}
\begin{figure}[h]
    \centering
    \includegraphics[width = 0.8\linewidth, height = 1.5cm]{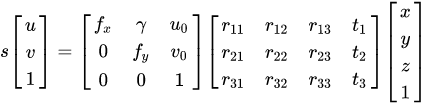}
    \captionsetup{justification=centering} \caption{\textit{PnP formula with extrinsic and camera intrinsic transformation matrices to project LiDAR points onto the image frame \protect\cite{b9}.}}
    \label{fig:eqn}
\end{figure}
\vspace{5pt}

\begin{figure}[h]
    \centering
    \includegraphics[width = \linewidth, height = 2.5cm]{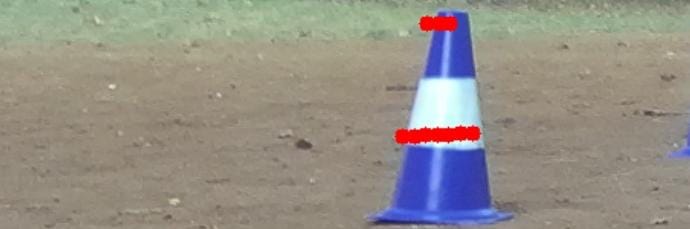}
    \captionsetup{justification=centering} \caption{\textit{Due to inherent errors in transformation (re-projection errors), few points lie outside the cone. Taking the average of the depth of each point reduces this error}}
    \label{fig:lidar}
\end{figure}
\vspace{6pt}

\noindent
\textbf{Point Cloud to Image Projection}
The 3D coordinates in the LiDAR frame are transformed to the camera frame, as the homogenous transformation matrix between the two is known due to fixed \& rigidly mounted sensors.  Then, the 2D point correspondence is deduced using the camera intrinsic parameters, which are also known. Thus, all cone points are transformed from 3D (x,y,z) to 2D (u,v) and the points which lie inside bounding boxes are used to find the depth of that bounding box. (Fig. \ref{fig:eqn} shows the formula for this transformation). The average depth of all projected points inside the bounding box is returned in order to keep the estimation robust to outliers. Cones (bounding boxes) for which no transformed point was found is passed onto the vision pipeline for depth estimation. The LiDAR projected points on the image is shown in Fig. \ref{fig:lidar}.

\subsection{Monocular Vision Pipeline}
\label{mono}

\paragraph{Using Bounding Box Dimensions}
A low latency approach for monocular depth estimation is to exploit the inverse relation of depth and the size of an object in an image. A curve fit for depth as a function of bounding box height for each cone is obtained. Self-collected data for approximately 230 cones placed at various angles, orientations, distances and illumination were used for this purpose. The method fails when there are edge cases like half-visible cones or fallen-over cones ("bad cones") because the incorrect bounding box height of such cones gives a false sense of depth.

\paragraph{Using Perspective Transformations} \label{pnpnpnpnp}

An alternate monocular vision pipeline involves the use of the Perspective n-Point (PnP) algorithm to estimate the relative pose of cones. This approach shares the bounding box detection by YOLOv5 and keypoint detection by RekTNet with the stereo vision approach (Section \ref{stereo}). However, there is only an image from a single camera used. These keypoints are used by the PnP algorithm to determine the homogeneous transformation matrix between an object’s local reference frame and the camera reference frame.

Given:
\vspace{-2mm}
\begin{itemize}[noitemsep]
  \item A set of \textit{n} 3D points $[x,y,z,1]^T$ (here, \textit{n = 7}) in the \textit{object} frame (known because the dimensions of the competition cones are known beforehand),
  \item The corresponding 2D image projections $[u,v,1]^T$(here, the 7 keypoints in \textit{image} frame), and,
  \item The calibrated intrinsic camera parameters (focal length, principle points, and distortion)
  \end{itemize}

The PnP algorithm estimates the pose of the camera ($[R|t]$) relative to the object frame, using the relation given by Fig. \ref{fig:eqn}. The translation vector $[t]$ gives the location of the cone relative to the car. This method had been implemented using OpenCV's solvePnP method for testing but is not used due to relatively high error.

\subsection{Stereo Vision Pipeline}
\label{stereo}

\begin{table*}[]
\centering
\resizebox{\textwidth}{!}{%
\begin{tabular}{|c|cc|c|ccc|c|}
\hline
\multirow{2}{*}{Pipeline} &
  \multicolumn{2}{c|}{\multirow{2}{*}{Method}} &
  \multirow{2}{*}{Avg. Error (\%)} &
  \multicolumn{3}{c|}{Percentage of cones with error} &
  \multirow{2}{*}{Comments} \\ \cline{5-7}
 &
  \multicolumn{2}{c|}{} &
   &
  \multicolumn{1}{c|}{\textless 5\%} &
  \multicolumn{1}{c|}{\textless 10\%} &
  \textless 20\% &
   \\ \hline
\multirow{4}{*}{Stereo-vision} &
  \multicolumn{1}{c|}{\multirow{2}{*}{Full bounding box SIFT}} &
  Top 1 keypoint &
  12.32 &
  \multicolumn{1}{c|}{73.17} &
  \multicolumn{1}{c|}{88.69} &
  92.23 &
  \multirow{4}{*}{Avg. calculated with outliers included} \\ \cline{3-7}
 &
  \multicolumn{1}{c|}{} &
  Top 2 keypoints &
  21 &
  \multicolumn{1}{c|}{69.4} &
  \multicolumn{1}{c|}{84.03} &
  90.46 &
   \\ \cline{2-7}
 &
  \multicolumn{1}{c|}{\multirow{2}{*}{Slender bounding box SIFT}} &
  Top 1 keypoint &
  6.39 &
  \multicolumn{1}{c|}{73.78} &
  \multicolumn{1}{c|}{88.98} &
  93.61 &
   \\ \cline{3-7}
 &
  \multicolumn{1}{c|}{} &
  Top 2 keypoints &
  6.89 &
  \multicolumn{1}{c|}{64.31} &
  \multicolumn{1}{c|}{82.15} &
  93.61 &
   \\ \hline
Mono-vision &
  \multicolumn{1}{c|}{Perspective n Point} &
  7 keypoints &
  9.14 &
  \multicolumn{1}{c|}{41.4} &
  \multicolumn{1}{c|}{64.77} &
  90.54 &
  Avg. calculated with Outliers excluded \\ \hline
Mono-vision &
  \multicolumn{1}{c|}{Depth using BB height} &
  Depth = 0.498*h\textasciicircum{}(-0.954) &
  4.49 &
  \multicolumn{1}{c|}{77.25} &
  \multicolumn{1}{c|}{92.71} &
  98.28 &
  Curve fit obtained using 241 cones \\ \hline
Mono - Stereo Fusion &
  \multicolumn{1}{c|}{Depth using BB height and Stereo pipeline for bad cones} &
  Depth = 0.498*h\textasciicircum{}(-0.954) &
  3.38 &
  \multicolumn{1}{c|}{78.96} &
  \multicolumn{1}{c|}{94.42} &
  100 &
  "Bad cones" are fallen / half visible ones \\ \hline

LiDAR - Camera Fusion &
  \multicolumn{1}{c|}{Using LiDAR depth and YOLOv5 class} &
  Tranformation equation used &
  0.85 &
  \multicolumn{1}{c|}{100} &
  \multicolumn{1}{c|}{100} &
  100 &
  No cone has \textgreater{}5\% error \\ \hline
\end{tabular}%
}

\captionsetup{justification=centering} \caption{Relative error in depth estimation for various pipelines obtained on real-world static testing}
\label{table:1}
\end{table*}

\begin{figure}[h]
    \centering
    \includegraphics[width = \linewidth]{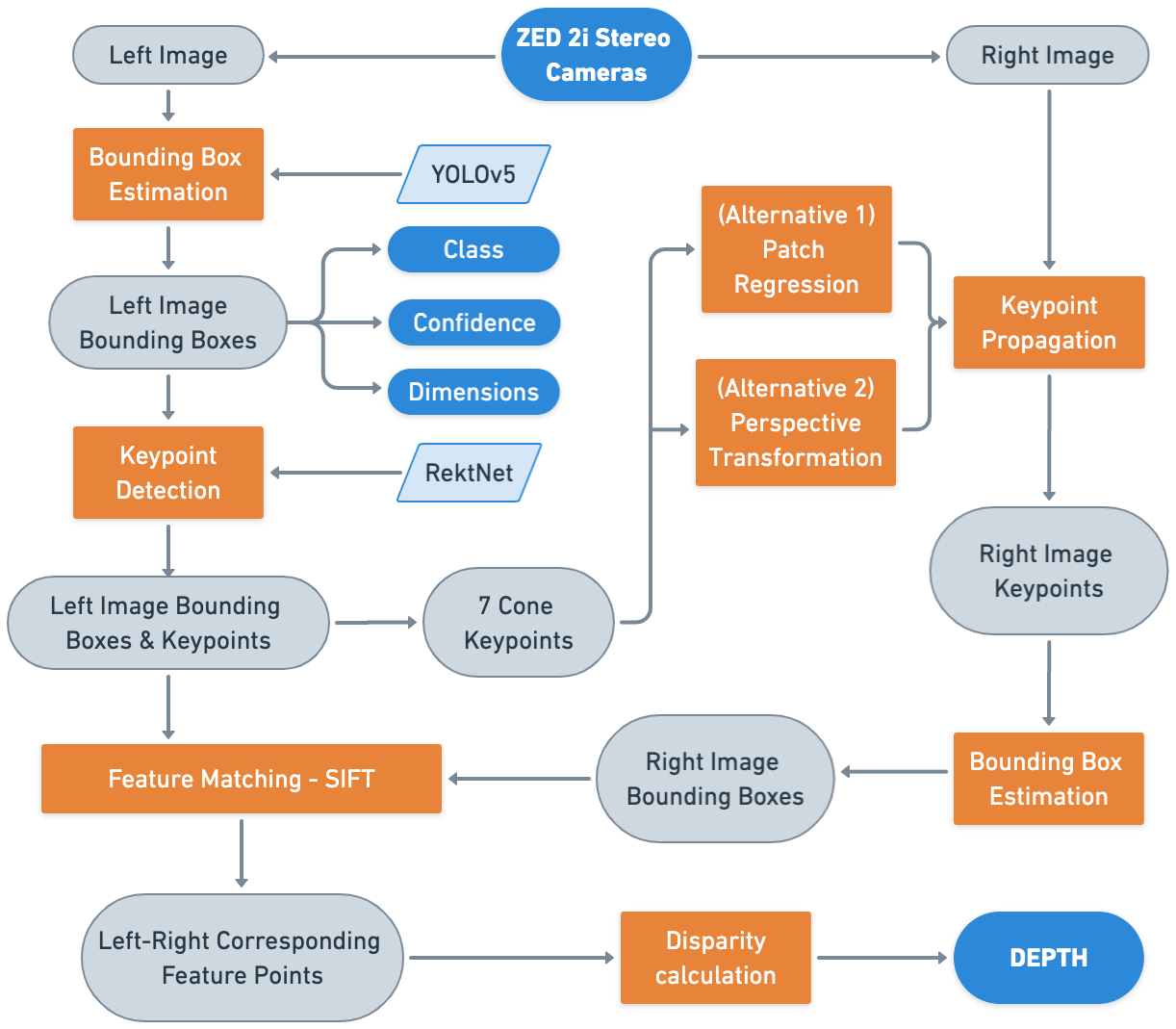}
    \captionsetup{justification=centering} \caption{\textit{Stereo vision depth estimation}}
    \label{fig:perception1}
\end{figure}

Stereo imaging commonly utilizes the parallax principle to estimate depth, which exploits the inverse relation between parallax shift (disparity) and the depth of an object. This pipeline implements the same principle sparsely, i.e. depth is calculated for each landmark, not for each pixel. Fig. \ref{fig:perception1} depicts the working of the stereo vision pipeline, which consists of the following major processes - i) Keypoint Detection and ii) PnP Transform and Feature Matching, followed by disparity calculation. These processes are detailed below.

\vspace{2mm}

\paragraph{Keypoint Detection}

\begin{figure}[h]
    \centering
    \includegraphics[width = 0.6\linewidth]{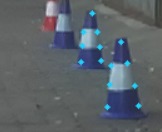}
    \captionsetup{justification=centering} \caption{\textit{7 Keypoints detected using RektNet on different cones}}
    \label{fig:keypoint}
\end{figure}

\textit{Seven salient keypoints} on every detected cone in the left image are found by the model RekTNet \cite{b4}. ReKTNet processes each bounding box separately to extract these keypoints as shown in Fig. \ref{fig:keypoint}. IIT Bombay Racing Driverless annotated and trained on an additional ~1000 images. The model works effectively even for non-standard FS Driverless cones.
\vspace{5pt}

\begin{figure}[h]
    \centering
    \includegraphics[width = 0.3\linewidth , height =85pt]{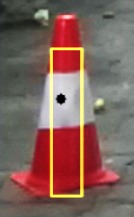}
    \captionsetup{justification=centering} \caption{\textit{"Slender" cone section utilized for SIFT feature matching}}
    \label{fig:BB}
\end{figure}

\paragraph{PnP Transform \& Feature Matching}
The homogeneous transformation matrix between the left camera frame and cone frame is found using the Perspective n-Point transformation (Fig. \ref{fig:eqn}).
Since the transformation matrix between the left camera and the right camera (stereo baseline shift) is also known, an estimate of the keypoints in the right image is obtainable by projecting the 3D coordinates of the cone frame onto the right image, providing the left-right corresponding keypoints. These keypoints are used to draw bounding boxes on the right image. Scale Invariant Feature Transform (SIFT) \cite{b14} is run on the obtained bounding boxes of left and right images to detect better features and match them with lesser error in disparity. Errors in the camera intrinsic matrix can lead to cascading of errors in PnP bounding box propagation. SIFT was run on a cropped part of the cones ("Slender" bounding boxes) to eliminate points outside the cone of interest - such as the ground and other overlapping cones in the bounding box (Fig. \ref{fig:BB}). These matched points are used to calculate disparity, and subsequently, the depth of a cone.

\vspace{1mm}

\subsection{Perception Results \& Performance}
Table \ref{table:1} tabulates the errors obtained in estimating depth using the various pipelines on static data collected using the stereo camera and LiDAR. The average error reduction for "Slender" bounding box SIFT is attributed to the removal of points that lie outside the cone. Top (best matched) 1 or 2 pair of keypoints by SIFT have been used for depth calculation to test the trade-off between redundancy to outliers and accuracy. It is observed that a single best keypoint gives the most accurate results.
The results highlight the highly accurate results of the LiDAR - Camera fusion pipeline, which justifies its use as the primary pipeline. Monocular being more accurate than Stereo, is used for all cones which are upright and fully visible. Monocular is however a rudimentary approach which requires a curve-fit to be calculated every time the camera placement and orientation changes. Stereo-vision proves to be the most robust of all pipelines with the capability of running on all cones detected and is thus used as a last resort for such edge cases at the slight cost of its accuracy and speed.
"Outliers" are cones with $>20\%$ error in depth estimation.

\section{Simultaneous Localisation and Mapping}
The SLAM subsystem is responsible for creating an environment using discrete cones (range, bearing and colour information) as landmarks while simultaneously tracking the car's position within the map, both in the global (ground-fixed) frame. The approach was to minimize mean-squared error in localization, achieve robust data association, \& low latency output to PPC (described below).

\subsection{EKF SLAM}

\begin{figure}[H]
    \centering
    \includegraphics[width = \linewidth]{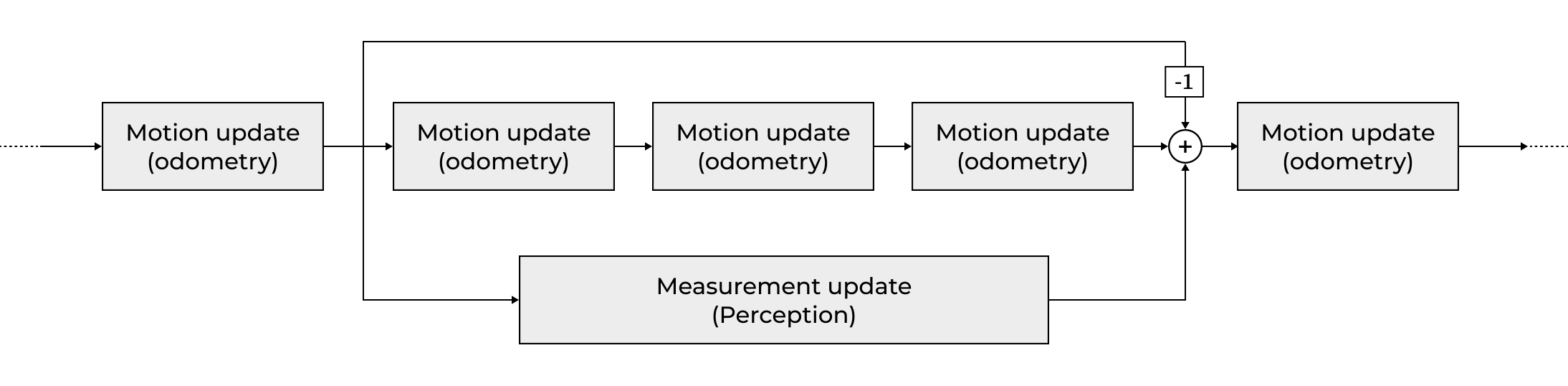}
    \captionsetup{justification=centering} \caption{\textit{Parallel structure of EKF SLAM; The difference in the current predicted car pose and the car pose when Perception data for measurement update  was taken is added to the measurement update pose estimate to account for time lag \protect\cite{b10}.}}
    \label{fig:motion}
\end{figure}

The Extended Kalman Filter (EKF) SLAM algorithm, with its real-time capability enabled by a modified architecture, proves adequate for feature-based mapping on competition tracks. It involves two principal stages: the Motion Update and the Measurement Update. Motion update uses odometry (IMU and wheel encoder) data to update the state of the car, while the measurement update
step acquires cone location and colour information from the perception subsystem to correct the motion update estimate through data
association, a step that critically impacts the system’s performance.
Motion update and Measurement update were parallelized as opposed to used in series. This allowed the motion update to continue predicting while executing the measurement update step in the background, thus reducing latency in data published to PPC. Fig. \ref{fig:motion} shows this parallel flow of motion and measurement update.

\paragraph{Motion Update} Uses odometry data to estimate the cars position, heading and velocity.
 The new car pose is adjusted for the difference in the initial pose to account for the delay made since the measurement was received. As a result, this approach provides a brief pose that the subsequent subsystem (i.e., Path Planning) can utilize for proper functionality even while Measurement Update is running in the background.

\paragraph{Measurement Update}
Corrects motion update using data from perception. It is computationally expensive, primarily due to the data association process which involves matching new measurements to existing map data and updating landmarks based on these new inputs. This process has a quadratic complexity $O(n^2)$, where n represents the number of landmarks \cite{b10}. Data association becomes particularly challenging with camera data, where errors can lead to incorrect pairing of cones.

\begin{figure}[H]
    \centering
    \includegraphics[width = 0.9\linewidth]{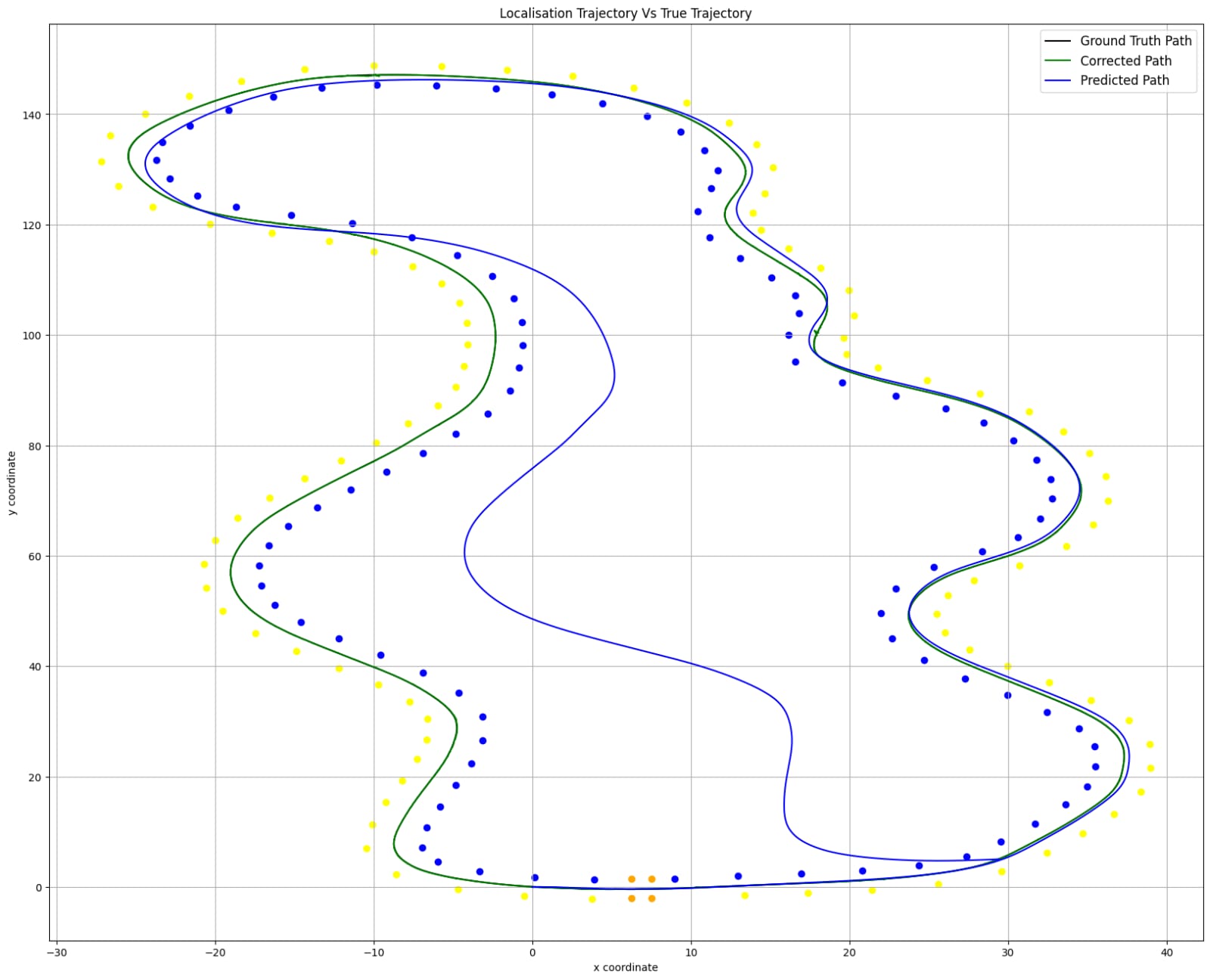}
    \captionsetup{justification=centering} \caption{\textit{The green path (corrected with measurement update) performs much better than the blue (only motion update) path}}
    \label{fig:localization}
\end{figure}

\begin{figure}[H]
    \centering
    \includegraphics[width = 0.9\linewidth]{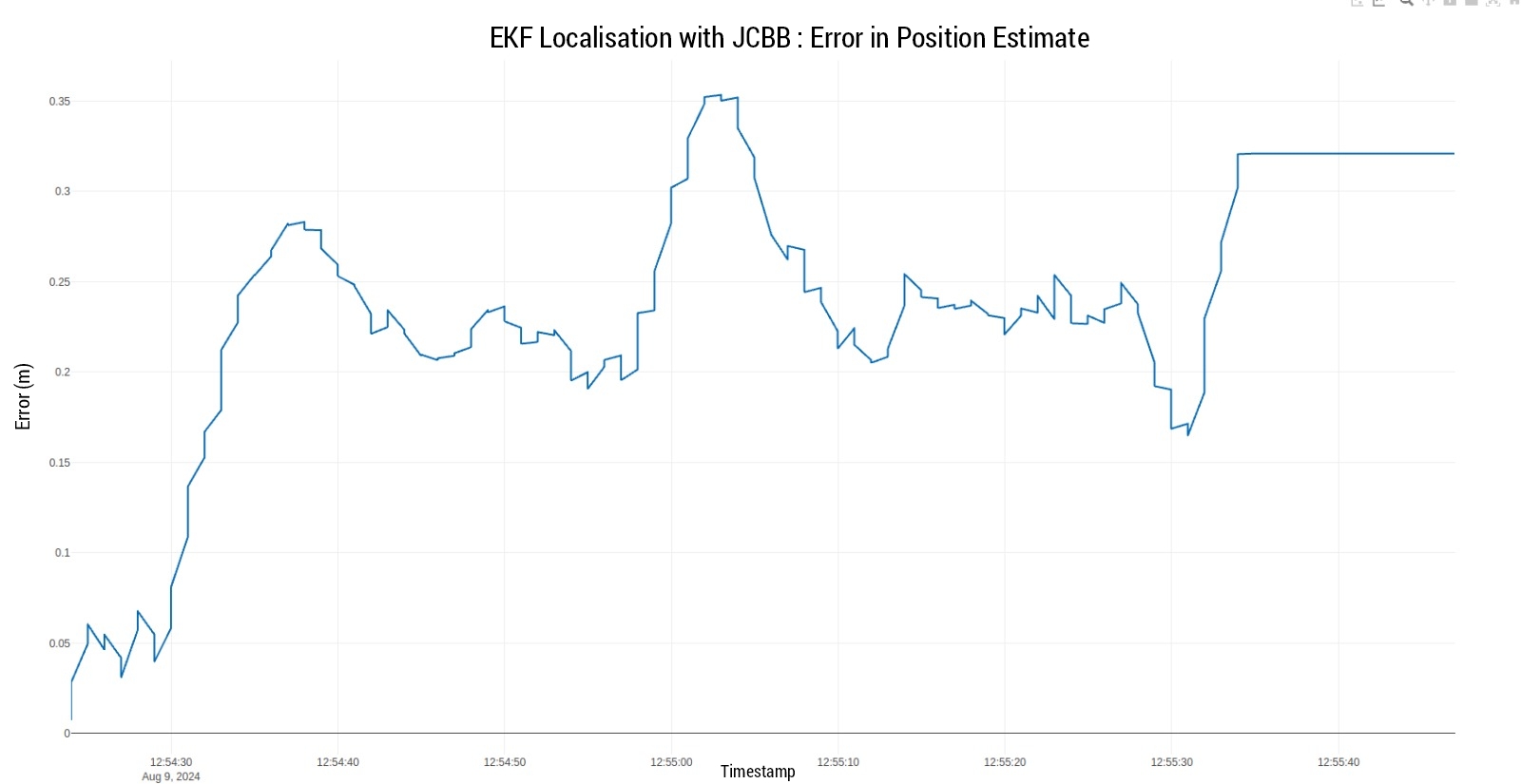}
    \captionsetup{justification=centering} \caption{\textit{EKF localization with JCBB data association: error in position estimate on Edinburgh University Formula Student (EUFS) simulator}}
    \label{fig:jcbb}
\end{figure}

\paragraph{Data Association} Cones that are seen for the first time are distinguished from those previously seen. New estimated cone positions are matched to existing map data and the map is updated. 
Currently , the Nearest Neighbour (NN) algorithm is used which identifies cones based on Euclidean distance. There is no covariance or sensor error taken into consideration while calculating the distances. Noise could lead to less accurate data association with multiple counting, double matching, not recognising a cone etc. To improve data association, data association algorithms like JCBB have been tested. Joint Compatibility Branch and Bound (JCBB)  algorithm explores all possible interpretations by systematically examining the tree structure. It aims to find the interpretation that includes the most pairings where the data points are jointly compatible and are non null.
The NN algorithm has been tested on localization only and the result is shown in Fig. \ref{fig:localization}. JCBB results are depicted in Fig. \ref{fig:jcbb}.

\subsection{Mobile Robot Programming Toolkit (MRPT)}
\begin{figure}[h]
    \centering
    \includegraphics[width = 0.6\linewidth]{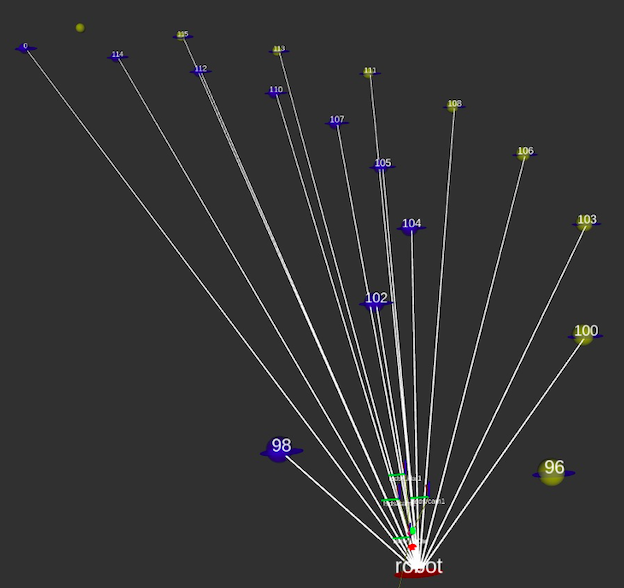}
    \captionsetup{justification=centering} \caption{\textit{Example of MRPT SLAM: White lines - Cones detected, Blue and Red Ellipses - Positional Covariance}}
    \label{fig:mrpt1}
\end{figure}

\begin{figure}[h]
    \centering
    \includegraphics[width = \linewidth]{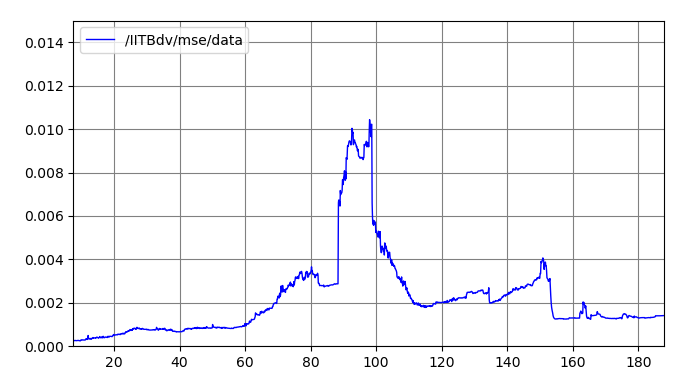}
    \captionsetup{justification=centering} \caption{\textit{Mean square error (MSE) for Observations vs No. of Cones Observed}}
    \label{fig:mrpt}
\end{figure}

The project also tested the SLAM and Localization nodes of the MRPT library \cite{b5}, which can process data from multiple sensors, including LiDARs and cameras. Fig. \ref{fig:mrpt1} depicts MRPT during simulation on Formula Student Driverless Simulator. The feature map published by the library contains easily processable information on the detected cones.  A custom implementation of MRPT was created to store additional landmark information such as colour which was not possible directly in the MRPT library used in the previous year, to the best of the authors' knowledge. 
Fig. \ref{fig:mrpt} depicts the mean square error of localized cones with actual positions compared with the number of cones observed. Two dips observed at ~100s and ~150s are due to the re-observation of old landmarks increasing accuracy suddenly.

\vspace{1mm}

\section{Path Planning and Controls}

Map data from SLAM is utilized to first plan the best suited trajectory and then provide control input to various actuators (steering, brakes and throttle) to achieve the desired trajectory. Mission status is also monitored to start and stop the car when desired, or to enter emergency mode. Cone data can also directly be used from Perception without the use of SLAM in events like the Acceleration Run where the map is fixed and does not feature any turns.

\begin{figure}[h]
    \centering
    \includegraphics[width = 0.7\linewidth]{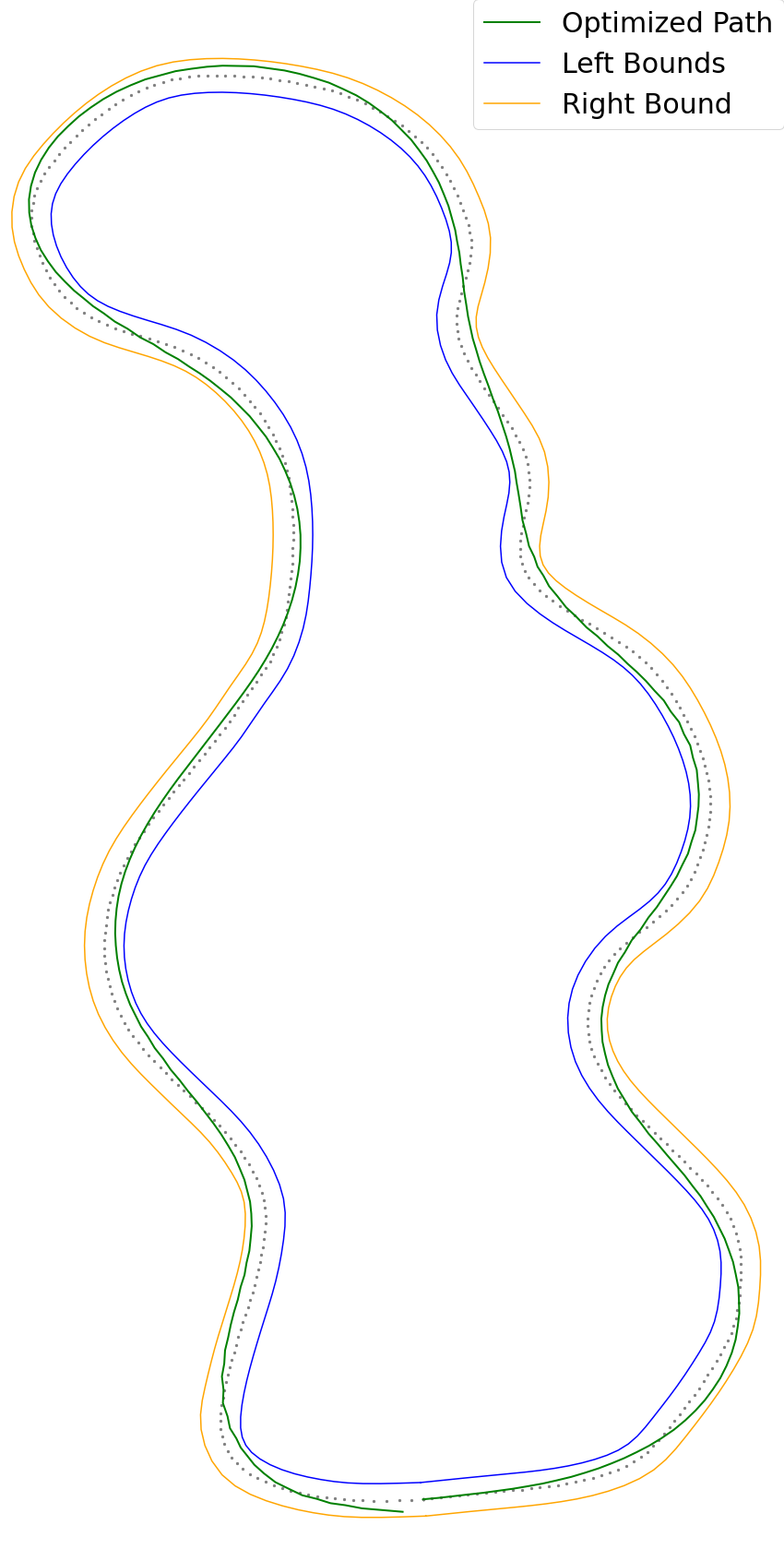}
    \captionsetup{justification=centering} \caption{\textit{Green: minimum curvature path, blue \& yellow: track boundaries}}
    \label{fig:pure_pursuit}
\end{figure}

\subsection{Path Planning}

In previous FS seasons, the planned path was formed by interpolating the track's midpoints obtained by cone-pairs chosen on the basis of a distance threshold. This has now been replaced by another algorithm - Delaunay Triangulation, which is a more robust version with significantly more desired way-points. 
 A faster path is also obtained by minimizing the curvature of the path. It logically follows that to minimize the lap time, the average speed on the track needs to be increased, and the distance travelled needs to be lesser. The latter constraint can be relaxed due to a relatively small track width compared to the lap distance. Thus, minimizing the curvature of the path was considered a good enough approximation to the minimization of lap time, since a reduction in path curvature leads to a better average speed due to a higher allowance for cornering velocities during turning. This is in accordance with the concept of a 'racing line'. This is depicted in Fig. \ref{fig:pure_pursuit}.

\begin{figure}[h]
    \centering
    \includegraphics[width = \linewidth]{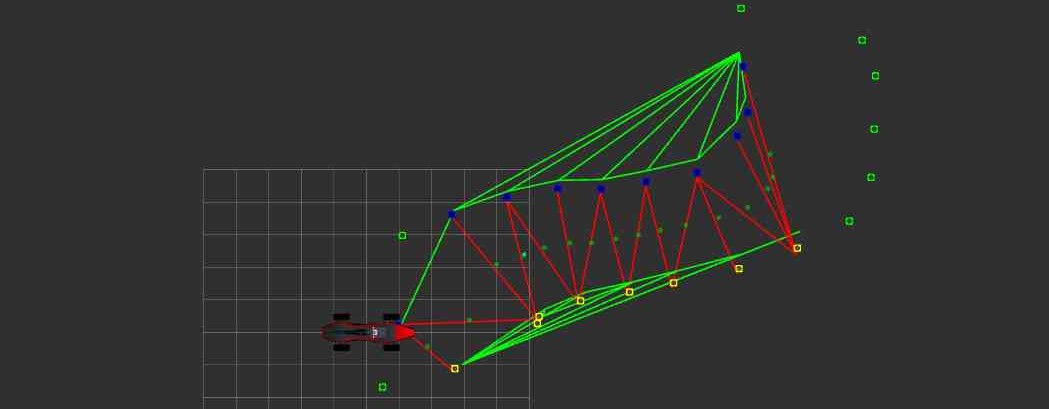}
    \captionsetup{justification=centering} \caption{\textit{Red lines: Triangulation within the track used for planning}}
    \label{fig:delaunay}
\end{figure}

The Delaunay algorithm is utilized to generate waypoints for the initial lap. It is formed by constructing triangulations in which the circumcircle associated with the three vertices of any triangle does not encompass any additional points. This is achieved through the creation of a Voronoi diagram. Once the midpoints are obtained, those situated between two cones of identical color are eliminated. Efforts are also made to exclude midpoints connecting cones from disparate track sections. Delaunay triangulation gives approximately double the number of waypoints when compared to a traditional midpoint trajectory and is hence used. Simulated working of this algorithm is depicted in Fig. \ref{fig:delaunay}.

\vspace{6pt}

\vspace{2pt}
\subsection{Control algorithms}
Stanley Control is an automatic steering method that uses the kinetic bicycle model. It is implemented by calculating the steering commands based on the deviation from a reference path (cross-track error) and the heading error of the vehicle. The Pure Pursuit controller is an automatic non linear steering method for
wheeled mobile robots. It involves computing the angular velocity control input that moves the vehicle/robot
from its actual position to reach a look-ahead point in front of
the vehicle/robot on the planned path. These path tracking algorithms enable the lateral control of the vehicle. Further, a PID controller is used to maintain a constant velocity. The distance between this look-ahead point and the robot is
proportional to its velocity. This is recursively done at each time step
to get a smooth steering control. The mean absolute cross track error reduced from 0.33 to 0.27 from Stanley control to Pure Pursuit control on the Formula Student Driverless Simulator.
\subsection{Path Planning for the Skidpad Event}
\begin{figure}[h]
    \centering
    \includegraphics[width = \linewidth]{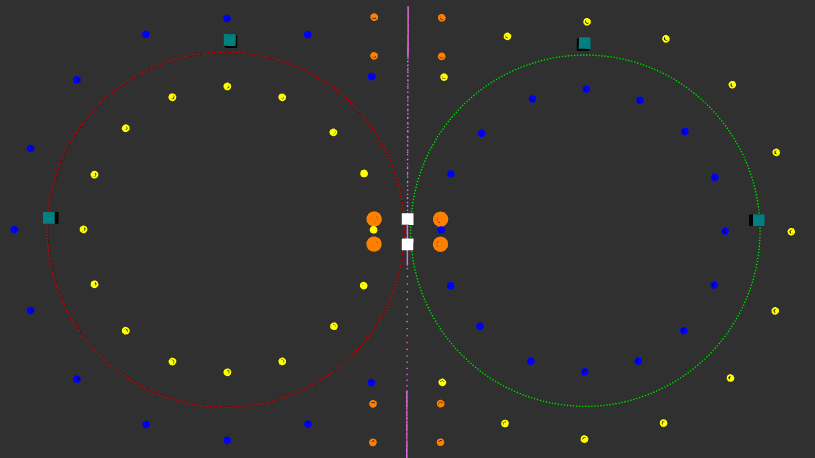}
    \captionsetup{justification=centering} \caption{\textit{Path planning for Skidpad}}
    \label{skidpad}
\end{figure}

A predetermined track is followed in the Skidpad event where the exact locations of the cones are given. The path is divided into 3 parts(red, green \& magenta), one straight line, and two circles on either side of the line (Fig. \ref{skidpad}) The paths selected are switched by using triggers in the center of the map (shown as White cubes in the figure), similar to that of a state machine. When the car is moving along any of the circles, it passes through two triggers (teal cubes in figure) and flags are set. A counter of each circle gets incremented when both the flags of that circle are set and then the flags are reset. The car switches its path when the counter is equal to 2 and it passes through the white triggers(cubes) in the center of the track. The Skidpad event has been completed in a time of 2 mins and 30 seconds on the Formula Student Driverless Simulator (FSDS).

\vspace{6pt}

\section{System Integration, Simulation Software and Scaled Testing Robot}
\label{hardware}

System Integration majorly deals with creating a ROS framework for the autonomous system. ROS communication protocol is utilized. 
 An intermediate node between camera node and perception node has been built to continuously receive images from the camera (at a higher frequency) and publish only when required to the perception node (as soon as a pipeline is done running). A time synchronizer is used to pass information only if data from two different sensors is obtained within a threshold time. A CAN interface node monitors the state of the vehicle and stores important event and operational data. Different subsystem algorithms are modularized into nodes and packages to keep debugging and testing simple and extensible.

\paragraph{Hardware Selection:}

\begin{itemize}[noitemsep]
    \item \textbf{Edge Compute Unit (NVIDIA Jetson AGX Orin):} Considered for its high computational power, low latency YOLOv5 benchmarks (used since the YOLOv5 algorithm is one bottleneck in the pipeline), TensorRT support and portable compact nature.
    
    \item \textbf{2x Stereo Camera (ZED 2i Stereo Camera):} It has a variable resolution and it also provides in-built depth estimation with easy ROS integration. The depth data from this camera was assumed the ground truth for testing in some cases.
    
    \item \textbf{LiDAR (Velodyne Puck VLP16):}  Commonly utilized LiDAR unit by FS-AI teams, it fit all requirements perfectly and was acquired as a sponsored unit.

    \item \textbf{IMU (SBG Systems Ellipse N GNSS/INS):} Extremely accurate and reliable odometry, also acquired as a sponsored unit.
\end{itemize}

\begin{figure}[H]
    \centering
    \includegraphics[width = \linewidth]{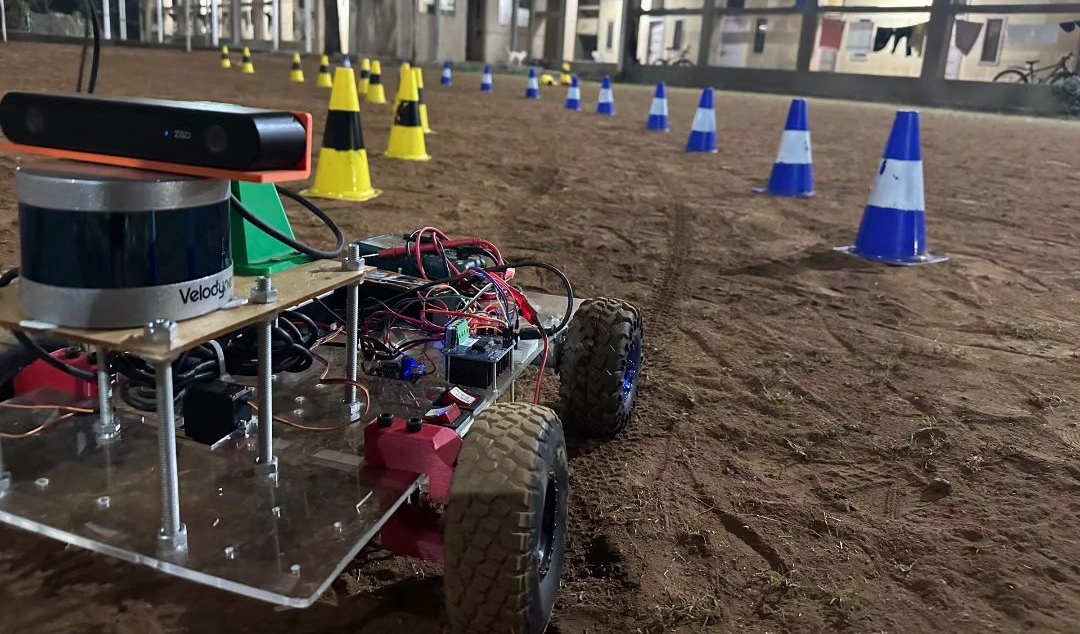}
    \captionsetup{justification=centering} \caption{\textit{Prototype robot for testing with a single stereo camera mounted}}
    \label{bot}
\end{figure}

\begin{figure}[H]
    \centering
    \includegraphics[width = \linewidth]{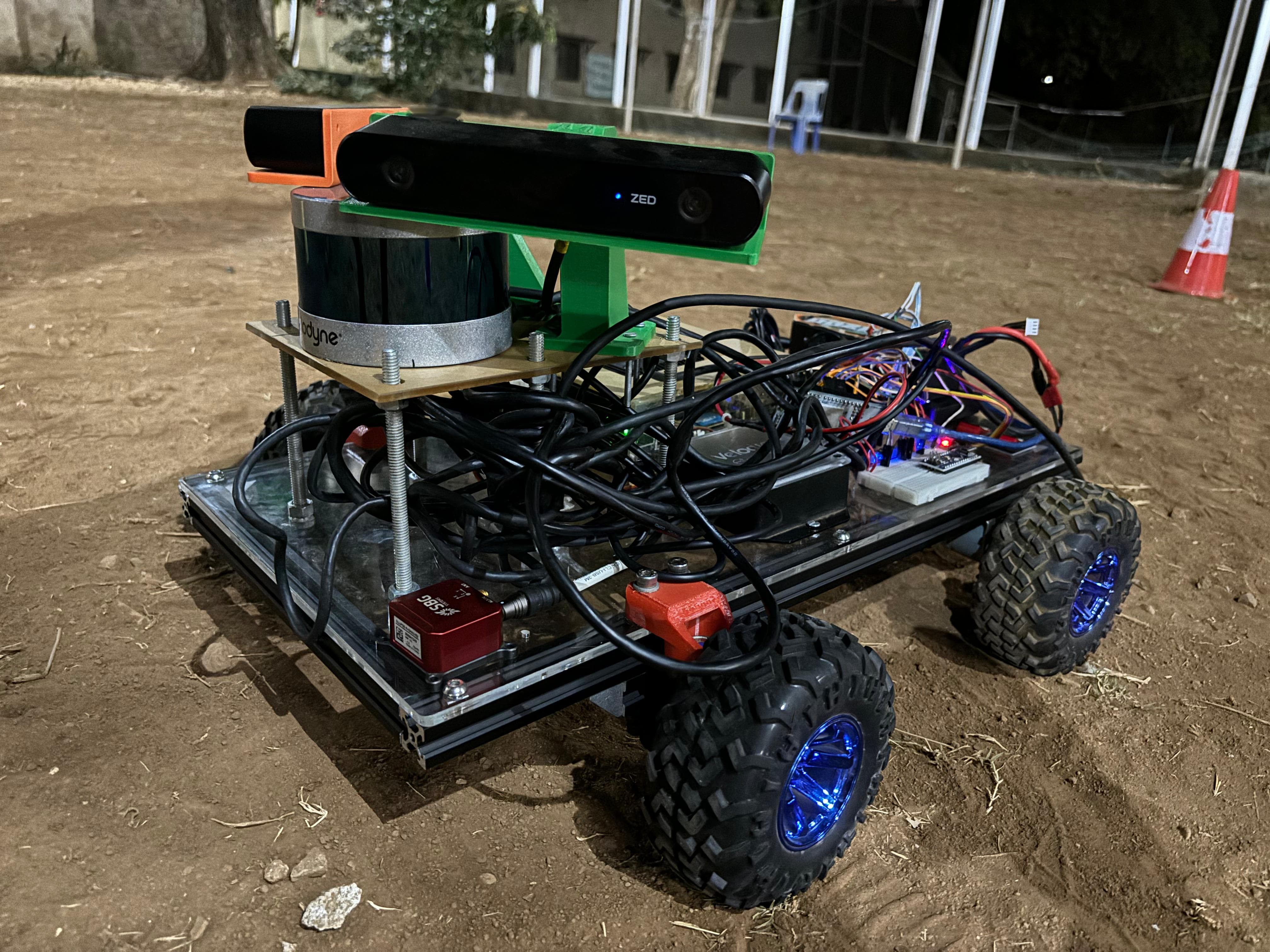}
    \captionsetup{justification=centering} \caption{\textit{Prototype robot for testing with two stereo cameras mounted}}
    \label{robot}
\end{figure}

Rigorous testing has been performed on dynamic simulations as well as static images.
 Edinburgh University Formula Student (EUFS) Simulator
 \cite{b7} and Formula Student Driverless Simulator (FSDS) \cite{b8} are used for simulations in all subsystems. A scaled robot was designed and built to test in close accordance with real life noise. It allows for gauging the performance of individual sub-systems when integrated and implemented in a real environment. Three iterations of the bot were built, with improvements in electronics and steering control. Extensive dynamic testing has been carried out in varying weather and visibility conditions. Water was also sprayed on the track to simulate cases of lack of friction and slipping. The algorithms have been found reliable through testing on bot with zero cones hit at slow speed for a single random lap. Fig. \ref{bot} shows the robot with a forward facing camera mount and Fig. \ref{robot} shows the robot with two side facing stereo cameras for a better horizontal field of view.

\section{Conclusion}
The autonomous software system stack comprising Perception, Simultaneous Localization and Mapping and Path Planning and controls has been developed for Formula Student AI, UK and the various algorithms have been detailed. Results for various approaches tried have been compared. These comparisons have justified the use of a parallel 3-tier approach for Perception with a LiDAR based pipeline, Stereo-vision and Monocular-vision being used together. The use of a sensor fusion based pipeline for perception has been justified with accuracy, latency and robustness to edge cases as key considerations. EKF SLAM and Mobile Robot Programming Toolkit (MRPT) have been used along with Stanley and Pure Pursuit based control for motion planning and controls. All subsystems have been tested on the self made robot and simulators with results described in above sections. This has proven crucial in ensuring reliability of the pipeline. Further work on simulating real life edge cases to make the system more robust is underway.
\vspace{4pt}

\vspace{7pt}

 \section*{Acknowledgments}
 On behalf of IIT Bombay Racing, we would like to thank our Faculty Advisor Prof. Sandeep Anand for his support and guidance over the years and Prof. Vivek Sangwan for insights on robot development, documentation and the support provided in publishing. Furthermore, we would like to thank IIT Bombay and the Institute Technical Council for funding the technical project. Furthermore, we would like to thank our driverless hardware sponsors: (Previously) Velodyne and SBG Systems without the support of which, the project could not have been realized. Furthermore, we would like to thank the Alumni from IIT Bombay Racing team whose donations and funding have helped us take part in the competition and source necessary hardware. Finally, we would like to thank all the student team members whose continuous efforts have led to the development of this project, including members from the mechanical, electrical, operations and marketing divisions of the team.

\end{document}